\documentclass{article}

\usepackage[preprint]{neurips_2025}

\usepackage[utf8]{inputenc} 
\usepackage[T1]{fontenc}    
\usepackage{hyperref}       
\usepackage{url}            
\usepackage{booktabs}       
\usepackage{amsfonts}       
\usepackage{nicefrac}       
\usepackage{microtype}      
\usepackage{xcolor}         
\usepackage{amsmath}
\usepackage{listings}
\usepackage{svg}
\usepackage{placeins}
\usepackage{adjustbox}

\title{Reflect, Retry, Reward:\\Self-Improving LLMs via Reinforcement Learning}

\author{%
  \textbf{Shelly Bensal}\thanks{Equal contribution} \quad
  \textbf{Umar Jamil}\footnotemark[1] \quad
  \textbf{Christopher Bryant} \quad \textbf{Melisa Russak}\\
  \textbf{Kiran Kamble} \quad \textbf{Dmytro Mozolevskyi} \quad \textbf{Muayad Ali} \quad \textbf{Waseem AlShikh} \\
  Writer, Inc. \\
  \texttt{\{shelly, ..., waseem\}@writer.com}
}

\begin{document}

\maketitle

\begin{abstract}
We explore a method for improving the performance of large language models through self-reflection and reinforcement learning. By incentivizing the model to generate better self-reflections when it answers incorrectly, we demonstrate that a model’s ability to solve complex, verifiable tasks can be enhanced even when generating synthetic data is infeasible and only binary feedback is available. Our framework operates in two stages: first, upon failing a given task, the model generates a self-reflective commentary analyzing its previous attempt; second, the model is given another attempt at the task with the self-reflection in context. If the subsequent attempt succeeds, the tokens generated during the self-reflection phase are rewarded. Our experimental results show substantial performance gains across a variety of model architectures, as high as 34.7\% improvement at math equation writing and 18.1\% improvement at function calling. Notably, smaller fine-tuned models (1.5 billion to 7 billion parameters) outperform models in the same family that are 10 times larger. Our novel paradigm is thus an exciting pathway to more useful and reliable language models that can self-improve on challenging tasks with limited external feedback.

\end{abstract}

\section{Introduction}

Large language models (LLMs) have demonstrated impressive capabilities across a wide range of natural language processing tasks \citep{zhao2025surveylargelanguagemodels}, as well as in mathematics \citep{ahn-etal-2024-large}, coding \citep{jiang2024surveylargelanguagemodels}, and reasoning \citep{huang-chang-2023-towards}. Despite these advancements however, models still have blind spots, and there is no guarantee that a model that succeeds at one task will succeed at another, even if the task is of a similar type \citep{asher-etal-2023-limits,huckle2025easyproblems}. The most direct way to address this problem is to retrain or fine-tune a model on data that represents the failed task, however this may not be possible if no such dataset exists. Furthermore, if the largest state-of-the-art models also struggle to complete the task, we similarly cannot use them to generate synthetic training data \citep{liu2024best}.

An alternative solution is to prompt the model to explain its reasoning or self-reflect on why it failed. For example, the popular Chain-of-Thought (CoT) paradigm \citep{wei2022chain} showed that models performed significantly better at arithmetic, commonsense, and reasoning tasks if they were prompted to show their reasoning in addition to simply providing a response. Self-reflection operates on a similar principle, in that if we can detect when a LLM provides an incorrect response, we can prompt it to reflect on any flaws in its reasoning and perhaps try again \citep{ji-etal-2023-towards,renze2024self}. The main advantage of these approaches is that they do not require any additional training data, however their effectiveness is directly tied to the effectiveness of the reasoning/reflection prompt. 

In this paper, we investigate the extent to which LLMs can learn to generate better self-reflections in order to self-improve on downstream tasks. More specifically, if a model fails to complete a task on its first attempt, it generates a self-reflection which it uses to make a second attempt. If the model then succeeds on its second attempt, we use reinforcement learning (RL), specifically Group Relative Policy Optimization (GRPO) \citep{shao2024deepseekmathpushinglimitsmathematical}, to reward the tokens in the self-reflection, such that future self-reflections will be more effective. In this way, models can learn how to improve upon all kinds of tasks without requiring any task-specific data; they instead just optimize how to reflect on mistakes.

Our main contribution is thus a novel methodology for training a model to generate better self-reflections to improve on challenging tasks in a task-agnostic way. Crucially, this method only requires a binary success/failure signal from a response verifier, which makes it well-suited to tasks where success can be easily verified. To demonstrate the efficacy of our approach, we carry out experiments on both the APIGen function calling dataset \citep{liu2024apigen} and the Countdown equation task introduced by \citet{tinyzero}.

\section{Related Work}

\subsection{Self-Reflection}

\textbf{Self-reflection in LLMs} Self-reflection, also referred to as introspection, is a metaprompting strategy in which a language model analyzes its own reasoning in order to identify and correct potential mistakes. This paradigm has gained momentum in large language model (LLM) research as a means to boost multi-step reasoning and problem-solving performance, especially in domains such as arithmetic, commonsense reasoning, and question answering \citep{wei2022chain, madaan2023self, renze2024self, shinn2023reflexion}. Typically, self-reflection involves generating an initial answer, producing natural language feedback to critique that answer, and then refining the response based on this critique. This process can be applied iteratively, often using the same model to both generate and evaluate solutions, and may include modules such as memory buffers or explicit meta-instruction guides \citep{liu2025instructofreflectionenhancinglargelanguage, wu2025rethinkingchainofthoughtperspectiveselftraining}.

\textbf{Approaches and Limitations} The methodology for self-reflection in LLMs varies along several axes. Some methods apply self-correction only to failed or low-confidence queries, while others use it for every response; feedback can be provided in the form of scalar scores, external annotations, or natural language, and may be generated by humans, external models, or the LLM itself \citep{Bai2022TrainingAH, peng2023checkfactstryagain, yang-etal-2022-re3, pan2025lemmalearningerrorsmathematical}. While prompting LLMs to self-reflect does improve accuracy in many settings, recent work has shown that the effectiveness depends strongly on the context: challenges include the inability to reliably identify self-errors without ground-truth oracles, diminishing returns from repeated reflection, and risks of performance deterioration for easier prompts or high-performing base models \citep{huang2024large, zhang-etal-2024-self-contrast, kim2023language}. In particular, self-reflection is most effective when initial accuracy is low, question difficulty is high, and external verification is available. Conversely, LLMs may sometimes fail to recognize their own mistakes but can still benefit from external feedback when such supervision exists \citep{pan2025lemmalearningerrorsmathematical, shinn2023reflexion}. 

\textbf{Training-Based Methods} Recent directions focus on incorporating self-improvement capabilities during model training, either by fine-tuning on self-correction trajectories or by formulating the process as a multi-turn reinforcement learning problem \citep{kumar2024traininglanguagemodelsselfcorrect, qu2024recursiveintrospectionteachinglanguage, wu2025rethinkingchainofthoughtperspectiveselftraining}. These training-based methods suggest that leveraging the model's own critiques during learning yields persistent improvements—even when no test-time self-reflection is performed. However, these approaches typically rely on larger teacher models for data generation or supervision, which can be seen as a form of knowledge distillation \citep{hinton2015distillingknowledgeneuralnetwork}.


\textbf{Our Approach} Building on insights from prior research, we propose correcting only failed cases identified by an external verifier, converting its binary feedback into self-reflective prompts, and training the model to use the self-reflection to succeed at the second attempt. This oracle-grounded conditional computation leverages training-time benefits to reduce test-time overhead and is guaranteed to improve or maintain performance, since corrections are applied only to initially incorrect examples. For training, we employ Group Relative Policy Optimization (GRPO), introduced in the next section. Notably, this approach bootstraps solely from the model's own outputs, without relying on external LLMs.

\subsection{Reinforcement Learning for Language Models}

\textbf{GRPO} Group Relative Policy Optimization (GRPO) is an outcome-based reinforcement learning method proposed to address the unique challenges faced when fine-tuning LLMs, such as those encountered in complex mathematical reasoning tasks \citep{shao2024deepseekmathpushinglimitsmathematical}. Unlike conventional approaches like Proximal Policy Optimization (PPO) \citep{schulman2017proximalpolicyoptimizationalgorithms}, GRPO dispenses with a separate value (critic) network and instead estimates advantages directly by comparing outcomes from a group of sampled completions. This makes GRPO particularly well-suited to settings where supervision is sparse and only available at the conclusion of a generation—for example, whether a completed math solution is correct. In such environments, the model must generate an entire sequence before receiving any feedback, typically in the form of a scalar reward reflecting the quality or correctness of the output.

\textbf{Our Approach} In this work, we adopt GRPO as the sole mechanism for reinforcement learning, without involving additional supervised fine-tuning stages. Recent research has demonstrated that modifying GRPO’s reward structure can effectively encourage models to persist through failure, for instance by rewarding retries after unsuccessful attempts, thereby promoting self-correction and robustness \citep{dao2025rezeroenhancingllmsearch}. GRPO has further shown promise in related domains requiring complex, outcome-supervised behaviors—including tool use and advanced mathematical problem solving—offering a flexible and efficient optimization strategy in diverse LLM applications \citep{qian2025toolrlrewardtoollearning, li2025torlscalingtoolintegratedrl}.

\section{Reflect, Retry, Reward}

\begin{figure}
    \centering
    \includegraphics[width=1.0\textwidth]{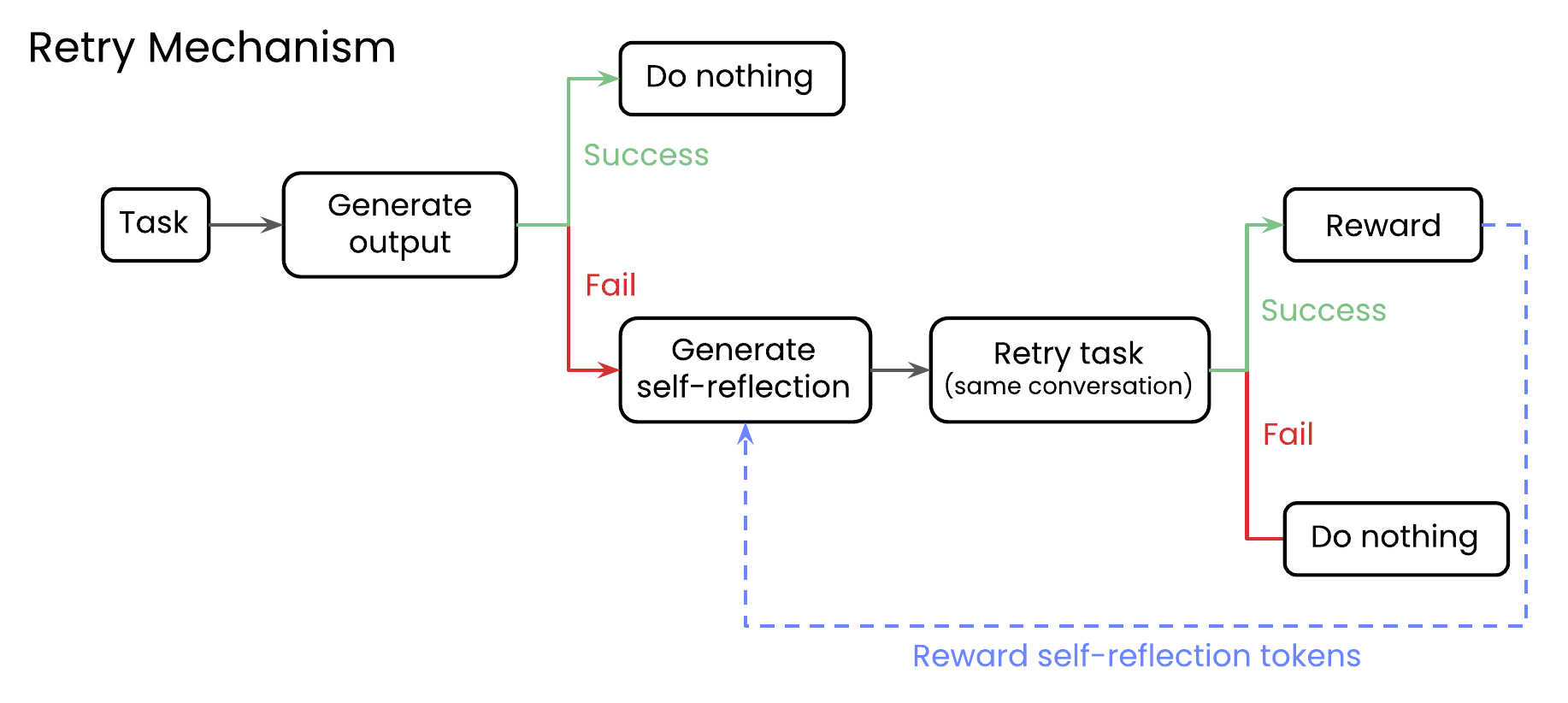}
    \caption{\textbf{Reflect, Retry, Reward Mechanism} The model is first prompted to complete a task based on a user query. If the initial response is correct, the process stops. If not, the model is prompted to generate a self-reflection on how to improve. The model then retries the same task, this time with its self-reflection included, and the new answer is evaluated. If the second attempt succeeds, the model learns that it generated an effective self-reflection.}
    \label{fig:figure_continue_conversation}
\end{figure}

Our novel Reflect, Retry, Reward methodology operates as follows, and is illustrated in \autoref{fig:figure_continue_conversation}.

First, a model is prompted to complete a task. If it succeeds, we do nothing as the model already meets our needs. If it fails however, we prompt it to generate a self-reflection on what might have gone wrong. Note that this presupposes a validator that automatically evaluates whether a response was a success or failure (binary). While it is sometimes possible to define a task-dependent validator that meets this criteria without ground-truth labels, such as in basic API function calling (Did the API call return a valid response?), mathematical equations (Does the equation evaluate to the target answer?), or code (Does the generated code execute?), some task types may require gold-standard target answers.  

Having generated a self-reflection, the model then makes a second attempt to complete the task, making use of the self-reflection in the conversation history. If it still fails, we do nothing; the self-reflection was insufficient to turn a failure into a success. If it succeeds however, we use GRPO to reward \textit{only the tokens that were generated in the self-reflection}. This is possible by setting the advantage terms for all other generated tokens to zero. We do this because we want the model to learn how to self-reflect more generally rather than specialize for a particular task. In other words, we do not reward the correct answer, we only reward the self-reflection.

\section{Experiments}

We demonstrate the effectiveness of our approach through experiments on two different tasks: function calling and math equations.

\subsection{Function Calling}

We use the APIGen dataset \citep{liu2024apigen} for our function calling experiments. APIGen is a dataset of 60,000 high quality function calls that consist of a user query (plain text), a list of possible tools that can answer that query plus their parameters (JSON) and the correctly formatted function call with the correct parameters and values (JSON). There are a total of 4,211 unique tools in the dataset, with an average of 2.3 parameters per tool, and each user query has an average of 2.8 tools to choose from (min 1, max 8). A model is only considered to be correct if it not only selects the right tool, but also generates the correct parameters and values. A sample datapoint with a choice of two different tools is shown below (formatted to be more human readable).

\begin{lstlisting}
USER QUERY:
Check if the Vimeo username 'john_doe_artist' is available.

TOOLS PROVIDED:
[{  "name": "vimeo", 
    "description": "Checks if a given Vimeo username is available using the Toolbench RapidAPI service.",
    "parameters": {"username": {"description": "The Vimeo username to check for availability.", "type": "str", "default": "username"}}
    }, 
{   "name": "get_user_pins", 
    "description": "Retrieves the Pinterest pins of a specified user.",
    "parameters": {"username": {"description": "The Pinterest username whose pins are to be fetched.", "type": "str", "default": "0869178429hau"}}
}]

CORRECT ANSWER:
[{"name": "vimeo", "arguments": {"username": "john_doe_artist"}}]
\end{lstlisting}

To preserve the integrity of our experiments, we only evaluate models that were released \textit{before} the APIGen dataset was released (June 2024). This ensures it is impossible that any of these models could have been trained on the dataset to obtain an unfair advantage. Specifically, we report results for Qwen2 (1.5B/7B Instruct) \citep{yang2024qwen2technicalreport}, Llama3.1 (8B Instruct) \citep{grattafiori2024llama3herdmodels}, and Phi3.5-mini Instruct \citep{abdin2024phi3technicalreporthighly}. We also report the vanilla performance of Qwen2-72B Instruct, Llama3.1-70B Instruct, and Writer's Palmyra X4 \citep{writer-palmyra-x4} as a baseline.

Since different model families also have different suggested tool-calling approaches, we tested different templates for each model family and ultimately chose the prompt formats that provided the strongest baselines. For our function calling validator, we require the model output to exactly match the correct answer in the dataset (i.e. based on the ground-truth labels). We used the following prompt to generate self-reflections for failed function calling attempts:

\begin{lstlisting}
You tried performing the task, but failed in generating the correct tool call. Reflect on what went wrong and write a short explanation that will help you do better next time.
\end{lstlisting}

\subsection{Countdown Math Equations}

We use the Countdown dataset introduced by the TinyZero project for our math equation experiments \citep{pan2025learningadaptiveparallelreasoning,tinyzero}. The Countdown dataset consists of 450k lists of 3-4 numbers along with a target number. The goal is to apply basic arithmetic operations to the numbers such that the equation evaluates to the target number. A model is only considered to be correct if it uses all the numbers once (in any order) and if the final equation successfully evaluates to the target number. A sample datapoint is shown below.  

\begin{lstlisting}
Using the numbers [4, 73, 4, 23], create an equation that equals 76. You can use basic arithmetic operations (+, -, *, /) and each number can only be used once.
\end{lstlisting}

As with function calling, to preserve the integrity of our experiments, we only evaluate models that were released or have a knowledge cutoff \textit{before} the Countdown dataset was made publicly available (January 2025). Specifically, we report results for Qwen2.5 (1.5B/3B/7B Instruct) \citep{qwen2025qwen25technicalreport}, Llama3.1 (8B Instruct), Llama3.2 (3B Instruct), and Writer's Palmyra 1.7B. We also report the vanilla performance of Qwen2.5-32B Instruct, Qwen2.5-72B Instruct, Llama3.1-70B Instruct, and Writer's Palmyra X4 \citep{writer-palmyra-x4} as a baseline. 

We once again tried several different prompt formats for each model family, and ultimately chose to use the format that provided the strongest baseline. For our math equation validator, we required the generated equation to match the target answer in the prompt (i.e. no need for ground-truth labels). We used the following prompt to generate self-reflections for failed Countdown math equations:

\begin{lstlisting}
You tried solving the problem and got the wrong answer. Reflect on what went wrong and write a short explanation that will help you do better next time.
\end{lstlisting}

\subsection{A Dataset of Failures}

For reasons of efficiency, and to facilitate a more intuitive analysis, we did \textit{not} train our models on the full function calling and math equation training sets, but instead opted to first create a dataset of failures for each task. More specifically, we prompted each model for each task to generate up to 64 responses (depending on model size) to each user query and preserved only those queries where the model \textit{failed} (based on each task-dependent verifier). We typically generated more responses for larger models because they failed less frequently than smaller models, and so would otherwise yield fewer training samples. To accelerate the rejection sampling process, we used vLLM \citep{kwon2023efficient} with prefix caching.

This approach has several advantages. First and foremost, it saves time because there is no point training our self-reflection model on queries it already handles successfully and hence cannot learn from. Second, by generating several responses per query, we make the data more robust; for example, if a base model generates a correct response to the same query 80\% of the time, we can still learn from the remaining 20\% since responses are not deterministic. Finally, by only having failure cases in our dataset, we can precisely determine how many samples the model needed to train on before it converged on the optimum self-reflection.

We must emphasize that we took this approach purely for reasons of efficiency and analysis, and it is otherwise functionally equivalent to learning from a real-world scenario where we receive both successful and failed responses.

\subsection{Multi-Step GRPO}

We used the TRL framework \citep{vonwerra2022trl} as a starting base to implement our multi-step GRPO algorithm (i.e., to learn from the second attempt after self-reflection). In particular, we extend the \texttt{GRPOTrainer} and alter its \texttt{\_prepare\_inputs} function to call a \texttt{second\_step} function that, given the completions generated by the \texttt{GRPOTrainer}, will perform another step of completion generations, without affecting the mask already computed by the \texttt{GRPOTrainer}. As we operate on the dataset of failures, prompting the model to generate its self-reflection commentary, the mask corresponds to the tokens of the self-reflection text. This way, we can perform as many secondary steps on the initial completions as necessary and only reward the tokens (through the mask generated by the \texttt{GRPOTrainer}) corresponding to the initial completions. The \texttt{second\_step} function also adds a data structure to the inputs sent to the reward function that helps in understanding the performance of the initial completion on the successive steps. This multi-step approach allows us to integrate any complex downstream reward mechanism instead of only rewarding the initial completions.

We trained our models on the respective failure datasets for up to 1,750 steps with an effective batch size of 256 failures (though in practice most models converged significantly faster) and we evaluated them at their convergence point. For example, the function calling experiment on Llama-3.1-8B Instruct required only 100 training steps and utilized less than 2,000 unique queries. Only one function calling experiment saw the entire dataset of 48,000 queries; the average across all function calling experiments was less than 25,000 unique queries. The most any math equation writing experiments used was less than 25,000 unique problems; the average of all math equation writing experiments was around 15,000 unique problems.

We used standard GRPO training parameters as described in the original DeepSeek implementation \citep{shao2024deepseekmathpushinglimitsmathematical}, and conducted some hyperparameter experimentation. In our final experiments, we set the KL divergence coefficient to 0.001, and used a learning rate of 5e-7 with a cosine annealing schedule and a warmup ratio of 0.03. To train each model, we used between 4 and 8 H100 GPUs. We limit our experiments to models between 1.5 billion and 8 billion parameters due to known computational efficiency and scalability concerns with GRPO \citep{zhang2025grpoleaddifficultyawarereinforcementlearning}.

In addition to the experimental results reported here, we also carried out experiments with some smaller models. We quickly discovered, however, that these models had a very limited capacity to answer accurately and self-reflect; e.g. Qwen2/Qwen2.5 0.5B Instruct and Llama3.2-1B Instruct. Similarly, while Microsoft’s Phi 3.5 mini model was able to handle function calling, it struggled significantly with equation writing. We do not report results for these models.

\section{Experimental Results}

\begin{table}[t]
  \centering
    \begin{tabular}{lrrrr}
    \toprule
     & \textbf{Vanilla} & \textbf{+ Reflection} & \textbf{Trained} & \textbf{+ Reflection} \\
    \textbf{APIGen} & \textbf{1st Try} & \textbf{2nd Try} & \textbf{1st Try} & \textbf{2nd Try} \\
    \midrule
    Qwen-2-1.5B Instruct & 32.6\% & 34.8\% & 48.6\% & 52.9\% \\
    Qwen-2-7B Instruct & 66.4\% & 69.4\% & 72.2\% & 77.3\% \\
    Llama-3.1-8B Instruct & 64.9\% & 70.9\% & 68.7\% & 74.9\% \\
    Phi-3.5-mini Instruct (3.8B) & 47.5\% & 50.2\% & 52.9\% & 56.0\% \\
    \midrule
    Qwen-2-72B Instruct& 73.7\% & 76.6\% & -     & - \\
    Llama-3.1-70B Instruct & 66.8\% & 76.9\% & -     & - \\
    Palmyra-X4 (73B) & 79.9\% & 83.5\% & -     & - \\
    \bottomrule&
    \end{tabular}%
  \caption{\textbf{APIGen Results} This table shows model performance in terms of accuracy on our APIGen test set (12,000 samples) both on the first and second attempt, and with/without our GRPO self-reflection training.}
  \label{tab:apigen_results}%
\end{table}%

Our main experimental results are shown in \autoref{tab:apigen_results} and \autoref{tab:countdown_results}. Specifically, \autoref{tab:apigen_results} shows model performance for each model's first and second attempts on the APIGen test set (12,000 samples) both before and after our multi-step GRPO training, while \autoref{tab:countdown_results} shows the same but for the Countdown test set (15,000 samples). 

In terms of APIGen, we first note that model size correlates perfectly with model performance after one attempt (as expected). We also note that performance increased by an average of 4.5\% after a second attempt using a self-reflection, which again is in line with previous work. We see the biggest increase after our GRPO training however, where although we only reward self-reflection tokens, almost all models are able to outperform even the two-attempt vanilla models after just a single attempt. We hypothesize this is because the self-reflection tokens help with model reasoning in general, so the model benefits even if it does not need to generate an explicit self-reflection. Nevertheless, self-reflection still helps after our training, and performance increases a further 4.7\% (on average) when models can self-reflect for their second attempt. Most strikingly, we observe that our Qwen-2-7B model after GRPO training is able to outperform a vanilla Qwen-2-72B model when both models are given two attempts, even though the latter model is 10x bigger than the first.  

\begin{table}[t]
  \centering
    \begin{tabular}{lrrrr}
    \toprule
     & \textbf{Vanilla} & \textbf{+ Reflection} & \textbf{Trained} & \textbf{+ Reflection} \\
    \textbf{Countdown} & \textbf{1st Try} & \textbf{2nd Try} & \textbf{1st Try} & \textbf{2nd Try} \\
    \midrule   
    Qwen-2.5-1.5B Instruct & 6.0\% & 10.2\% & 34.9\% & 45.0\% \\
    Qwen-2.5-3B Instruct & 18.8\% & 29.0\% & 33.9\% & 47.3\% \\
    Qwen-2.5-7B Instruct & 31.7\% & 38.0\% & 41.6\% & 50.3\% \\
    Llama-3.1-8B Instruct & 2.2\% & 4.6\% & 8.8\% & 17.8\% \\
    Llama-3.2-3B Instruct & 2.1\% & 3.0\% & 8.8\% & 13.8\% \\
    Palmyra 1.7B & 26.8\% & 31.8\% & 33.3\% & 38.6\% \\ 
    \midrule
    Qwen-2.5-32B Instruct & 38.6\% & 45.1\% & -     & - \\
    Qwen-2.5-72B Instruct & 45.2\% & 49.9\% & -     & - \\
    Llama-3.1-70B Instruct & 17.3\% & 25.5\% & -     & - \\
    Palmyra-X4 (73B) & 46.8\% & 51.6\% & -     & - \\
    \bottomrule&
    \end{tabular}%
  \caption{\textbf{Countdown Results} This table shows model performance in terms of accuracy on the Countdown test set (15,000 samples) both on the first and second attempt, and with/without our GRPO self-reflection training.}
  \label{tab:countdown_results}%
\end{table}%

In terms of Countdown, it is first worth noting that performance was lower across the board, and the vanilla Llama models in particular (both Llama-3.1 and Llama-3.2) really struggled to complete the task; for example, the Llama-3.1.70B model was outclassed by even the Qwen-2.5-3B model, which is more than 20x smaller. Otherwise, the pattern of improvement is similar to the APIGen experiments, albeit at a slightly higher magnitude: self-reflection increased performance by an average of 5.3\% and 8.6\% respectively before and after our GRPO training. We hypothesize that these larger gains come from the fact the models started from a lower baseline and hence had a greater opportunity to learn.

Ultimately, our findings not only reinforce previous work on the benefits of self-reflection, they also demonstrate how learning to optimize for self-reflection with GRPO can improve performance further still.

\subsection{Better Self-Reflections}

\begin{figure}
    \centering
    \includegraphics[width=1.0\textwidth]{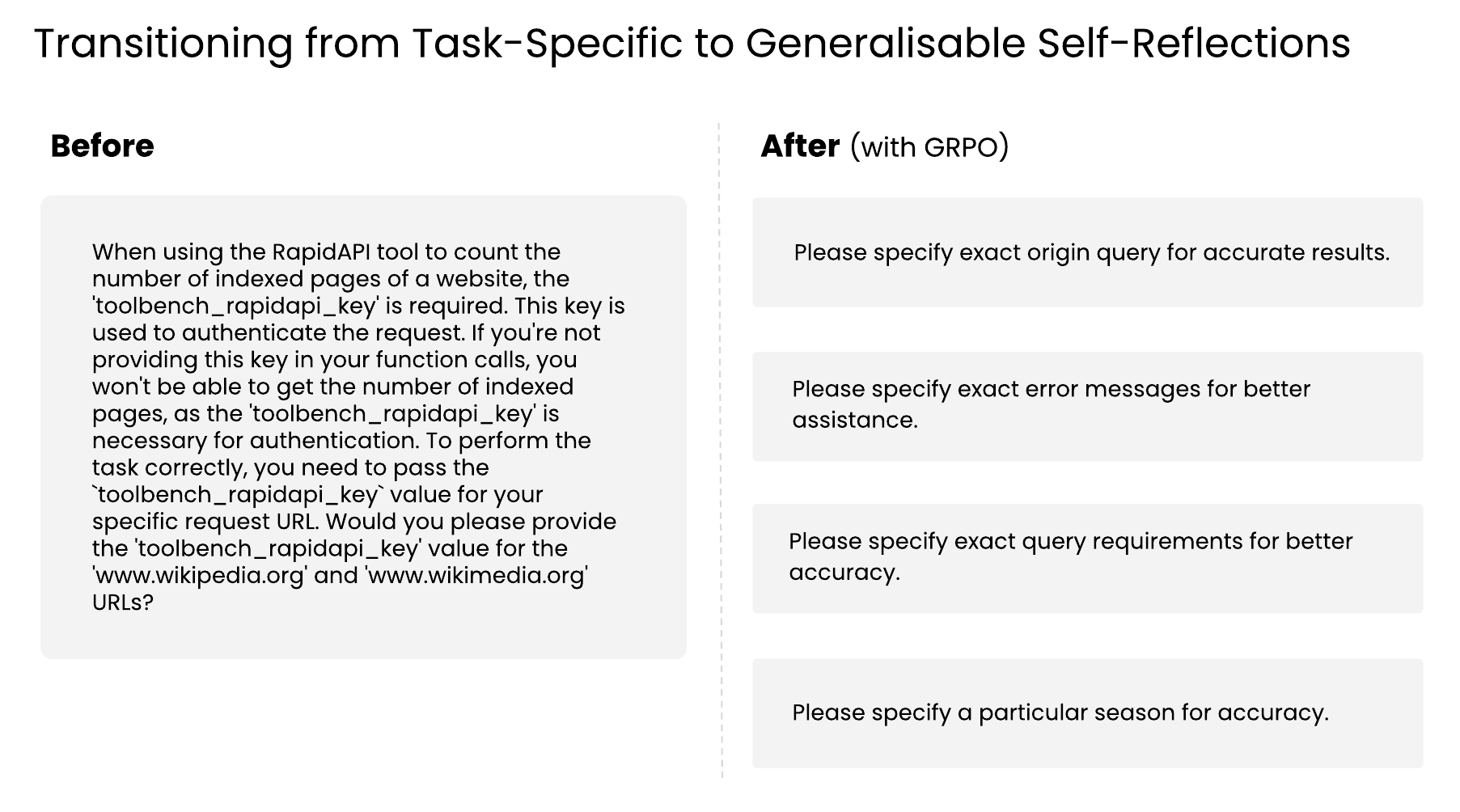}
    \caption{\textbf{Better Self-Reflections} We observe that reflections generated by vanilla models tend to be long, confusing, and redundant, whereas GRPO fine-tuned models produce much shorter, clearer, and more generalisable reflections.}
    \label{fig:figure_self_reflection}
\end{figure}

To provide an insight into how self-reflections improve after self-reflection training, we present a qualitative example of a self-reflection generated by a vanilla model alongside a self-reflection generated by the same model after GRPO training in \autoref{fig:figure_self_reflection}. It is immediately obvious that vanilla self-reflections are much longer, more verbose, and repetitive compared to the more concise, optimized self-reflections after training. While this intuitively makes sense -- humans likewise prefer short, simple instructions -- this finding contrasts with chain-of-thought-style outputs, which are believed to perform better precisely \textit{because} they are more verbose. We leave it as an open question as to when it may be more beneficial for a model to generate concise vs. verbose output.

\subsection{Low Catastrophic Forgetting}

\begin{table}
\centering
\begin{adjustbox}{width=\textwidth}
\begin{tabular}{lrr|rr|rr|rr}
\toprule
\textbf{Model} 
& \multicolumn{2}{c|}{\textbf{MMLU-Pro}} 
& \multicolumn{2}{c|}{\textbf{GSM8K}} 
& \multicolumn{2}{c|}{\textbf{HellaSwag}} 
& \multicolumn{2}{c}{\textbf{MATH}} \\
& \textbf{Vanilla} & \textbf{Trained} 
& \textbf{Vanilla} & \textbf{Trained} 
& \textbf{Vanilla} & \textbf{Trained} 
& \textbf{Vanilla} & \textbf{Trained} \\
\midrule
\multicolumn{5}{l}{\textit{Function calling models}} \\
Qwen2-1.5B Instruct & 22.1\% & 21.9\% & 59.6\% & 59.5\% & 66.0\% & 66.1\% & 21.2\% & 20.9\% \\
Qwen2-7B Instruct & 43.4\% & 42.8\% & 77.6\% & 77.0\% & 80.7\% & 80.5\% & 34.0\% & 33.7\% \\
Llama3.1-8B Instruct & 41.0\% & 40.9\% & 78.0\% & 77.7\% & 79.3\% & 79.3\% & 31.1\% & 31.5\% \\
Phi3.5-mini & 44.6\% & 44.5\% & 79.2\% & 79.7\% & 76.8\% & 76.8\% & 31.6\% & 31.0\% \\
\midrule
\multicolumn{5}{l}{\textit{Math equation writing models}} \\
Qwen2.5-1.5B Instruct & 30.9\% & 31.5\% & 59.4\% & 58.0\% & 68.3\% & 68.3\% & 25.5\% & 26.3\% \\
Qwen2.5-3B Instruct & 33.4\% & 32.7\% & 65.3\% & 65.1\% & 74.9\% & 74.9\% & 32.2\% & 31.6\% \\
Qwen2.5-7B Instruct & 44.8\% & 44.2\% & 84.2\% & 82.1\% & 80.5\% & 80.5\% & 42.3\% & 40.3\% \\
Llama3.1-8B Instruct & 41.0\% & 41.1\% & 78.0\% & 77.3\% & 79.3\% & 79.6\% & 31.1\% & 31.0\% \\
Llama3.2-3B Instruct & 31.5\% & 31.7\% & 65.8\% & 65.0\% & 70.4\% & 71.0\% & 28.3\% & 28.2\% \\
Palmyra 1.7B & 31.7\% & 31.7\% & 76.2\% & 76.3\% & 50.0\% & 49.9\% & 43.0\% & 43.0\% \\
\bottomrule&
\end{tabular}
\end{adjustbox}
\caption{\textbf{Catastrophic Forgetting Analysis} Comparison between vanilla and GRPO fine-tuned models on common LLM benchmarks shows that despite fine-tuning, we observe minimal catastrophic forgetting, with fine-tuned models maintaining strong performance on these standard benchmarks.}
\label{table_catastrophic_forgetting}
\end{table}

A common concern when fine-tuning models is catastrophic forgetting, i.e. when a model learns to specialize on one task at the expense of others \citep{Li2016LearningWF, 10.5555/3295222.3295393, kotha2024understandingcatastrophicforgettinglanguage}. Since our self-reflection training is designed to improve performance in a task agnostic way, we evaluate our models on several diverse benchmarks (MMLU-Pro \citep{wang2024mmluprorobustchallengingmultitask}, GSM8K \citep{cobbe2021trainingverifierssolvemath}, HellaSwag \citep{zellers2019hellaswagmachinereallyfinish}, and MATH \citep{hendrycksmath2021}) in order to assess their capacity for language understanding, mathematical problem solving, and commonsense reasoning both before and after self-reflection training. We do this using the common evaluation benchmark framework \texttt{lm-eval} \citep{eval-harness}. Our hypothesis is that performance should remain relatively unchanged, since we never optimize for a specific task, but instead optimize self-reflection reasoning in general.

We present our results in \autoref{table_catastrophic_forgetting}, and find that performance does indeed remain stable after self-inflection training. In most cases, there is less than 1\% degradation compared to the base model, and some models even improve; e.g. Qwen-2.5-1.5B performance increases by 0.6\% and 0.8\% respectively on MMLU-Pro and MATH after self-reflection training on the Countdown dataset. We treat this as evidence our approach is robust to catastrophic forgetting. 



\section{Conclusion}

In this paper, we have shown that it is possible to significantly improve LLM performance by training a model to improve at self-reflection rather than at a particular task. This indirect approach depends only on a validator that can detect whether a model response is correct or incorrect, and so is particularly well-suited to tasks where responses can be easily verified; e.g. whether JSON output is formatted correctly, whether generated code is actually executable, or whether all the constraints of an equation are satisfied. 

We demonstrated the efficacy of our approach through experiments on the APIGen function calling and Countdown math equation solving datasets, and found that models trained for self-reflection using GRPO improved performance by an average of 9.0\% on the function calling test set (12,000 samples) and 16.0\% on the Countdown match equation dataset (15,000 samples). We furthermore found that smaller self-reflection trained models could outperform larger untrained models on both tasks, despite their size difference; e.g. Qwen-2-7B Instruct (trained) outperformed Qwen2-72B Instruct (untrained) on function calling, and Qwen2.5-7B Instruct (trained) outperformed Qwen2.5-72B Instruct (untrained) on Countdown math equations. Our models were also robust to catastrophic forgetting.

Although we only trained models to improve at self-reflection, we found they also performed significantly better even when they did not need to self-reflect; i.e. they succeeded on the first attempt so there was no need to reflect and try again. We hypothesize that this is because by focusing on self-reflection rather than a particular task, models may have improved their reasoning skills more generally. In future work, we hope to investigate whether self-reflection training generalizes across different tasks.

\section{Limitations}

It may not always be straightforward to define a binary success/fail validator for every task. We developed our method with the view that labeled training data may be scarce, but recognize that ground-truth labels could be used as a validator if available. Alternatively, it may also be possible to use a larger model as a judge \citep{zheng2023llmjudge}. 


We also find that our approach does not work for all models and all tasks; the model must have some basic ability to perform the task, self-reflect, and learn in order for boosting self-correction ability to work. For example, Llama3.2-3B Instruct was unable to learn to self-correct on the function calling task.

\bibliographystyle{acl-natbib}
\bibliography{main}

\newpage

\appendix

\section{Prompt Templates}

For reproducibility and clarity, we provide details on the prompt templates used during training. To the best of our ability, we followed model provider recommendations for prompting. We iterated on prompts to achieve reasonable baselines for each model on each task. 

\subsection{Function Calling}

\paragraph{Qwen 2 models} Our prompting style for Qwen 2 models for function calling is as follows. First, we provide the following system prompt:

\begin{lstlisting}
You are a helpful assistant that can answer questions and help with tasks. 

# Tools

You may call one or more functions to assist with the user query.

You are provided with function signatures within <tools></tools> XML tags:
<tools>
{List of tools, each on a new line}
</tools>

For each function call, return a json object with function name and arguments within <tool_call></tool_call> XML tags:
<tool_call>
{\"name\": <function-name>, \"arguments\": <args-json-object>}
</tool_call>
\end{lstlisting}

This is followed by the user query from the dataset, as role user. The model then replies with its first attempt at the task. If the attempt is incorrect, we prompt for a self-reflection as follows:

\begin{lstlisting}
You tried performing the task, but failed in generating the correct tool call. Reflect on what went wrong and write a short explanation that will help you do better next time.
\end{lstlisting}

After the model generates a self-reflection, we again prompt with the system prompt and user query to set up the model for its second attempt at the task.

\paragraph{Llama 3.1 and Phi 3.5 models} We follow the recommended Llama 3.1 tool calling format. We found that Phi performs better following the Llama tool-calling template than the Qwen 2 template.  First, we provide the following system prompt: 

\begin{lstlisting}
When you receive a tool call response, use the output to format an answer to the original user question.

You are a helpful assistant with tool calling capabilities.
\end{lstlisting}

Then, as role user, we provide the tools and user query from the dataset as follows:

\begin{lstlisting}
Given the following functions, please respond with a JSON for a function call with its proper arguments that best answers the given prompt.

Respond in the format {\"name\": function name, \"parameters\": dictionary of argument name and its value}. Do not use variables.

{List of tools, each on a new line}

Question:
\end{lstlisting}

This is followed by the user query from the dataset, as role user. The model then replies with its first attempt at the task. We then prompt for a self-reflection:

\begin{lstlisting}
You tried performing the task, but failed in generating the correct tool call. Reflect on what went wrong and write a short explanation that will help you do better next time.
\end{lstlisting}

After the model generates a self-reflection, we prompt with just the user query to set up the model for its second attempt at the task.

\subsection{Countdown Math Equations}

We provide the following system prompt: 

\begin{lstlisting}
Please reason step by step, and put your final answer within \\boxed{}.
\end{lstlisting}

Then, as role user, we provide the main problem as follows:

\begin{lstlisting}
 Using the numbers {nums, in list format} create an equation that equals {target}. You can use basic arithmetic operations (+, -, *, /) and each number can only be used once.
 Please reason step by step, and put your final answer within \\boxed{}.
\end{lstlisting}

The model then replies with its first attempt at the task. Given a failure, we then prompt for a self-reflection:

\begin{lstlisting}
You tried solving the problem and got the wrong answer. Reflect on what went wrong and write a short explanation that will help you do better next time.
\end{lstlisting}

After the model generates a self-reflection, we repeat the user message from above to set the model up for its second attempt at the task: 

\begin{lstlisting}
 Using the numbers {nums, in list format} create an equation that equals {target}. You can use basic arithmetic operations (+, -, *, /) and each number can only be used once.
 Please reason step by step, and put your final answer within \\boxed{}.
\end{lstlisting}

\newpage

\begin{table}[t]
  \centering
    \begin{tabular}{lrrrr}
    \toprule
    \textbf{APIGen} & \textbf{Tool choice error} & \textbf{Parameter error} & \textbf{Format error} \\
    \midrule
    Vanilla Qwen-2-1.5B Instruct & 33.9\% & 25.3\% & 8.2\% \\
    Trained Qwen-2-1.5B Instruct & 21.3\% & 25.3\% & 4.8\% \\
    \midrule
    Vanilla Qwen-2-7B Instruct & 5.2\% & 27.2\% & 1.2\% \\
    Trained Qwen-2-7B Instruct & 3.5\% & 22.7\% & 1.6\%\\
    \midrule
    Vanilla Llama-3.1-8B Instruct & 4.6\% & 28.5\% & 2.0\% \\
    Trained Llama-3.1-8B Instruct & 3.7\% & 25.6\% & 2.0\% \\
    \midrule
    Vanilla Phi-3.5-mini Instruct (3.8B) & 18.8\% & 24.8\% & 8.9\% \\
    Trained Phi-3.5-mini Instruct (3.8B) & 17.6\% & 23.9\% & 5.7\% \\
    \bottomrule&
    \end{tabular}%
  \caption{\textbf{APIGen Error Analysis} This table categorises model errors on the first attempt at the task with and without GRPO self-reflection training (12,000 sample test set).}
  \label{tab:fc_error_analysis}%
\end{table}%

\begin{table}[t]
  \centering
    \begin{tabular}{lrrrr}
    \toprule
    \textbf{Countdown} & \textbf{Invalid equation} & \textbf{Wrong numbers} & \textbf{Missed target} \\
    \midrule
    Vanilla Qwen-2.5-1.5B Instruct & 5.9\% & 73.7\% & 14.4\% \\
    Trained Qwen-2.5-1.5B Instruct & 0.6\% & 34.3\% & 30.2\% \\
    \midrule
    Vanilla Qwen-2.5-3B Instruct & 4.9\% & 52.5\% & 23.9\% \\
    Trained Qwen-2.5-3B Instruct & 2.1\% & 33.6\% & 30.4\% \\
    \midrule
    Vanilla Qwen-2.5-7B Instruct & 3.8\% & 39.1\% & 25.4\% \\
    Trained Qwen-2.5-7B Instruct & 0.1\% & 55.6\% & 2.7\% \\
    \midrule
    Vanilla Llama-3.1-8B Instruct & 0.1\% & 97.1\% & 0.6\% \\
    Trained Llama-3.1-8B Instruct & 0.0\% & 90.4\% & 0.7\% \\
    \midrule
    Vanilla Llama-3.2-3B Instruct & 0.1\% & 97.1\% & 0.7\% \\
    Trained Llama-3.2-3B Instruct & 0.0\% & 90.4\% & 0.7\% \\
    \midrule
    Vanilla Palmyra 1.7B & 4.1\% & 57.5\% & 11.6\% \\
    Trained Palmyra 1.7B & 3.0\% & 44.6\% & 19.1\% \\
    \bottomrule&
    \end{tabular}%
  \caption{\textbf{Countdown Error Analysis} This table categorises model errors on the first attempt at the task with and without GRPO self-reflection training (15,000 sample test set).}
  \label{tab:countdown_error_analysis}%
\end{table}%

\section{Error Analysis}
We categorise the errors of our models before and after training in an attempt to better understand what types of errors models are prone to on these tasks, and what types of errors can be mitigated by self-reflection training. We look exclusively at errors made on the first attempt at the task (pass@1). 
\subsection{Function Calling}

For function calling, we categorise errors into three types: errors in tool choice, errors in parameter names or values, and errors in format. We consider parameter choice to be much more difficult than tool choice.

The two smallest models (Qwen-2-1.5B Instruct and Phi-3.5-mini Instruct) struggle significantly with tool choice without training, and don't improve much, if at all, on parameter values through training. Conversely, the larger models (7-8 billion parameters) are already quite good at tool choice without training, and training primarily seems to teach parameter selection.

\subsection{Math Countdown Equations}

For math countdown equations, we categorise errors into three types: an invalid equation (or one that uses disallowed characters), an equation that uses numbers outside of the provided ones (wrong numbers), and an equation that does not evaluate to the provided target (missed target). 

All models struggled primarily with outputting equations that used only allowed numbers. Training significantly decreased this error for all models except for Qwen-2.5-7B Instruct. Put another way, all models except the largest Qwen model primarily learned to use the correct numbers in the equation through training, even if it resulted in missing the target, whereas Qwen-2.5-7B Instruct learned to hit the target even if it meant using the wrong numbers.

\end{document}